\documentclass[lettersize,journal]{IEEEtran}
\usepackage[table]{xcolor}
\usepackage{amsmath,amsfonts}
\usepackage{algorithmic}
\usepackage{algorithm}
\usepackage{array}
\usepackage{textcomp}
\usepackage{stfloats}
\usepackage{url}
\usepackage{verbatim}
\usepackage{graphicx}
\usepackage{cite}
\usepackage{multirow}
\usepackage{booktabs}  
\usepackage{comment}
\usepackage{bm}
\usepackage{mathrsfs}
\usepackage{subfigure}

\usepackage[justification=centering]{caption}
\pdfpageattr{/Group << /S /Transparency /I true /CS /DeviceRGB>>}

\begin{document}

\title{Advancing Direct Training for Spiking Neural Networks with Circulate-Firing Neurons and Learnable Gradients}

\author{
    Feifan Zhou,
    Xiang Wei,
    Yang Liu,
    and Qiang Yu
\thanks{
    All authors are with the School of Artificial and Intelligence,
    Tianjin University, Tianjin 300072, China.
}
\thanks{
    Corresponding author: Qiang Yu
    (e-mail: yuqiang@tju.edu.cn).
}
}




\maketitle

\begin{abstract}
Spiking Neural Networks (SNNs) have emerged with promising energy-efficient property, yet a substantial performance gap persists compared to Artificial Neural Networks (ANNs). 
This gap stems from at least two key limitations: first, conventional spiking neurons offer limited information representation capacity, underutilizing the rich dynamics of membrane potentials; second, fixed surrogate gradient (SG) functions across time steps leads to imprecise gradient propagation, impeding effective direct training. 
To address these two challenges, we propose a new direct training algorithm with three core innovations: first, a circulate-firing spiking neuron model that enhances information representation capacity by leveraging membrane potentials more effectively; second, a time-step-wise learnable surrogate gradient function, enabling accurate gradient estimation during backpropagation; third, a positive-negative balanced loss function to achieve equilibrium between positive and negative membrane potentials and further boost SNN performance. 
Extensive experiments demonstrate that our methods achieve competitive performance across multiple datasets. 
Our methods can generalize seamlessly to advanced architectures of Transformer, consistently outperforming existing methods. 
Our work highlights the effectiveness of further harnessing intrinsic membrane dynamics of SNNs for performance improvement, and thus open a new avenue for advancing high-performance spiking neural architectures. 
\end{abstract}


\section{Introduction}
Artificial Neural Networks (ANNs) have achieved remarkable success across diverse domains, including image classification~\cite{he2016resnet}, natural language processing~\cite{annnlp}, speech recognition~\cite{dong2018speech}, and autonomous driving~\cite{drive}.
However, their impressive performance comes at the expense of significant energy consumption. 
As the third generation of neural networks \cite{1997Networks}, Spiking Neural Networks (SNNs) have emerged as a promising alternative due to their substantially higher energy efficiency. 
Unlike ANNs, which transmit information through continuous-valued activations, SNNs communicate using discrete binary spikes. Owing to their asynchronous and event-driven firing mechanism, SNNs can efficiently process and transmit spatio-temporal information with greatly reduced energy costs, making them particularly well-suited for deployment on neuromorphic hardware platforms \cite{pei2019hardware, debole2019truenorth}. 

Despite these advantages, training deep SNNs remains challenging. The discrete spiking mechanism is non-differentiable, hindering gradient backpropagation, and the conversion of rich membrane potentials into binary spikes leads to substantial information loss. As a result, a noticeable performance gap persists between SNNs and conventional ANNs.

To enable effective training of SNNs, researchers have proposed three main approaches: unsupervised learning, ANN-to-SNN conversion, and direct training.
Unsupervised learning methods update synaptic weights in SNNs based on biologically inspired rules, such as spike-timing-dependent plasticity (STDP).
STDP \cite{stdp} is an unsupervised learning mechanism that adjusts synaptic weights according to the temporal sequence and interval between the spike firings of pre-synaptic and post-synaptic neurons.
However, this learning paradigm is not well-suited for training deep spiking neural networks.
To address the issue of effectively training deep SNNs, two alternative strategies have been developed: the ANN-to-SNN conversion method \cite{Cao2015} and direct training based on backpropagation through time (BPTT) \cite{BPTT}.
The conversion method builds upon the theory that the firing rates of spiking neurons in SNNs can approximate the activation values of their ANN counterparts \cite{2020Constructing, qcfs, rmp-snn, li2021calibration, offset2023}.
This approach circumvents the challenge posed by non-differentiable spike activity by transferring weights from a pre-trained ANN to an SNN with an identical architecture.
Although converted SNNs can achieve nearly lossless performance compared with their original ANNs, they typically require a large number of time steps to attain competitive accuracy \cite{qcfs, li2021calibration, offset2023}.
Furthermore, such converted SNNs generally lack the ability to effectively process neuromorphic datasets.

An alternative approach is direct training, in which a surrogate gradient function is employed to approximate the ill-defined gradient of the spiking activation function during backpropagation, thereby enabling the application of the BPTT algorithm for end-to-end optimization of SNNs \cite{2019Surrogate, 2016snnbp}.
To further enhance training efficiency, researchers have proposed the spatio-temporal backpropagation (STBP) algorithm \cite{STBP}, which jointly considers spatial and temporal dependencies to facilitate high-performance supervised learning in SNNs.
Despite the recent success of direct training algorithms \cite{tdBN, sew, PLIF, MLF, imloss, rmp-loss, ternary, TAB, TET}, two critical challenges remain unresolved.
The first limitation lies in the restricted information representation capacity of the conventional leaky integrate-and-fire (LIF) spiking neuron model.
The second pertains to the propagation of inaccurate gradients during training.
Existing direct training methods rely on surrogate gradient functions to approximate true gradients. 
However, the membrane potentials of spiking neurons exhibit complex temporal dynamics.
In most prior works, a fixed and identical surrogate gradient function is applied across all time steps, which constrains the precision of gradient estimation and hinders effective optimization.
Addressing the above two challenges is therefore crucial for improving SNN performance and narrowing the performance gap between SNNs and ANNs.

In this work, we propose a novel direct training algorithm that enables high-performance SNN training and significantly reduces the performance gap between SNNs and ANNs. 
We validate the effectiveness of our approach over a wide range of different datasets, including static images datasets, neuromorphic datasets, text classification datasets, and tactile datasets. 
The main contributions of this work are summarized as follows:

\begin{itemize}
\item We propose circulate-firing (CF) spiking neuron model with enhanced information representation capacity through alleviating the information loss caused by the conversion from real-valued membrane potential to binary spike tensors. 

\item We introduce a novel time-step-wise learnable surrogate gradient function to propagate more accurate gradient during backpropagation for effective training of SNNs and a positive-negative balanced loss function to drive membrane potentials around resting levels, which boosts the performance of SNNs. 

\item 
We conduct extensive experiments over a wide range of tasks, including static datasets, neuromorphic datasets, text classification datasets and tactile datasets. 
Our algorithm outperforms existing state-of-the-art direct training approaches and generalizes seamlessly to Transformer-based SNN architectures, achieving consistent and substantial performance improvements over state-of-the-art baselines and demonstrating better robustness.  
\end{itemize}

The remainder of this paper is organized as follows. Section~\ref{related work} reviews related studies on spiking neural network training methods. Section~\ref{preliminaries} introduces the LIF spiking neuron model and direct training approach of SNNs. 
Section~\ref{methods} details our proposed direct training algorithm, including the CF spiking neuron model, the time-step-wise learnable surrogate gradient function, and the positive–negative balanced loss function. Section~\ref{experimental results} presents the experimental results and analysis. Finally, Section~\ref{conclusion} concludes the paper and discusses potential directions for future research.

\section{Related Work}
\label{related work}

\subsection{The Training Algorithms for SNNs}

Researchers have proposed several algorithms to train SNNs, including spike-timing-dependent plasticity (STDP)~\cite{stdp}, ANN-to-SNN conversion~\cite{Cao2015} and direct training~\cite{STBP}. 
STDP is a biologically inspired synaptic learning rule that modulates synaptic weights according to the precise temporal correlation between pre-synaptic and post-synaptic spike events~\cite{stdp}.
Although several STDP-based approaches have been proposed to enhance the learning capability of SNNs~\cite{goupy2024stdp, mozafari2018firststdp}, they are generally limited to shallow networks and remain unsuitable for training deep SNNs.

ANN-to-SNN conversion has become a dominant strategy for large-scale SNN training.
In this paradigm, a pre-trained ANN model is converted into an SNN by transferring its weights~\cite{Cao2015}.
To mitigate the performance degradation during conversion, various optimization techniques have been introduced, including soft-reset mechanisms~\cite{rmp-snn}, spike calibration~\cite{li2021calibration, offset2023}, modified spiking neuron models~\cite{2020Constructing}, and tailored activation functions~\cite{qcfs}.
Furthermore, several training-free SNN-based Transformer architectures have been proposed within this framework~\cite{you2024spikezipconversion, jiang2024spatioconversion, hwang2024spikedattentionconversion}.

Direct training, on the other hand, enables the optimization of SNN parameters directly through surrogate gradients to overcome the non-differentiability of spiking activity~\cite{rmp-loss}.
Spatio-temporal backpropagation (STBP) ~\cite{STBP} pioneered this line of work by jointly updating weights across spatial and temporal domains, though it was primarily limited to small-scale datasets.
To enable large-scale SNN training, threshold-dependent batch normalization (tdBN) and other normalization techniques have been proposed~\cite{tdBN, bntt, mpbn, tebn, TAB}, alongside architectural optimizations tailored for SNNs~\cite{sew, MLF}.
Additionally, a variety of improvements—including novel loss functions~\cite{imloss, rmp-loss, TET}, optimized surrogate gradient functions~\cite{imloss, lian2023learnable, li2024directly, adaptivesmooth}, and advanced neuron models~\cite{PLIF, MLF, LTMD, DIETSNN, yao2022glif, tc-lif, ternary, psn, lm-h, lm-ht}—have further enhanced direct training performance.
Recent works have also extended direct training to SNN-based Transformers~\cite{zhou2022spikformer, zhou2024qkformer}, and explored hybrid frameworks that combine ANN-to-SNN conversion with direct fine-tuning~\cite{DIETSNN, lm-h, lm-ht}.
Despite these advancements, direct training approaches still lag behind their ANN counterparts in accuracy and convergence efficiency.
\vspace{-0.5em}

\subsection{Advanced Spiking Neuron Model}

Spiking neurons constitute the fundamental computational units of SNNs, and improving neuron models is an essential approach to enhance network performance.
To address the limited information capacity of the conventional LIF neuron, several studies have proposed advanced neuron models that enable multi-spike generation~\cite{MLF, ternary, lm-ht} or introduce learnable neuronal parameters to model biological heterogeneity~\cite{PLIF, LTMD, ding2022biologically, DIETSNN, DEXAT}.
While these methods have contributed to improving representational capability and biological realism, fully exploiting the rich subthreshold membrane potential information remains an open challenge in SNN design. 

\subsection{Surrogate Gradient}

Surrogate gradient methods are widely used to approximate the non-differentiable spike function, enabling effective backpropagation in SNNs.
However, standard surrogate gradient functions suffer from gradient vanishing and gradient mismatch issues, which can hinder convergence and limit the performance of SNN models.
To address the gradient vanishing problem, several works have proposed learnable surrogate gradient functions that dynamically adjust the gradient shape during training, thereby improving both accuracy and latency~\cite{lian2023learnable, imloss}.
To alleviate gradient mismatch, hybrid methods combining true gradients with surrogate ones have been developed, allowing networks to adaptively approximate the loss landscape~\cite{li2024directly, adaptivesmooth}.
Moreover, most existing approaches apply identical surrogate gradient functions across all time steps, ignoring temporal dynamics in spiking behavior.
Exploring time-dependent learnable surrogate gradient functions thus represents a promising direction for enhancing temporal gradient propagation and overall SNN training efficiency. 

In summary, prior works have contributed significantly to advancing SNN research in terms of training algorithms, neuron model design, and surrogate gradient optimization.
However, existing methods still suffer from information loss during spike conversion, inaccurate gradient estimation, and imbalance in membrane potential dynamics.
These limitations motivate our work, which aims to design a more expressive spiking neuron model, a temporally adaptive surrogate gradient mechanism, and a balanced loss function to further improve the performance of SNNs. 

\section{PRELIMINARIES}
\label{preliminaries}

\subsection{Spiking Neuron Model and Direct Training of Spiking Neural Networks}

In this part, we briefly overview the LIF neuron model, which is one of the most commonly adopted spiking neuron models.
The iterative expression of the LIF models~\cite{Wu2019} can be simply described as follows: 

\begin{equation}
\label{lifcharging}
    I_{i}^{t, l}=\sum_{j=1}^{N^{l-1}} w_{i j}^{l} s_{j}^{t, l-1},
\end{equation}   

\begin{equation}
\label{lifupdate}
    u_{i}^{t, l} = k_{\tau}v_{i}^{t-1, l} + I_{i}^{t, l},
\end{equation}
\vspace{-2.5em}

\begin{equation}
\label{liffiring}
    s_{i}^{t, l}=H (u_{i}^{t, l} - \theta^{l}),
\end{equation}
\vspace{-2.5em}

\begin{equation}  
\label{lifreset}
         v_{i}^{t, l} =
         \left\{\begin{array}{ll}
u_{i}^{t, l} (1 - s_{i}^{t, l}) +  s_{i}^{t, l}u_{\mathrm{reset}}, & \mathrm{hard \enspace reset},\\
u_{i}^{t, l} - s_{i}^{t, l}\theta^l, & \mathrm{soft \enspace reset}.
\end{array}\right.
\end{equation} 

Eq.(\ref{lifcharging}), Eq.(\ref{lifupdate}) and Eq.(\ref{liffiring}) represent the charging, membrane potential updating and firing process of LIF neuron, where $I$ denotes the weighted input; $k_\tau$ represents the decay factor; $u$ and $v$ respectively denote the membrane potential before and after the reset process. $H(\cdot)$ denotes the heaveside function; $\theta^{l}$ denotes the threshold of the $l$-th LIF neuron model. 
Eq.(\ref{lifreset}) describes the expression of two different reset mechanisms: hard reset (membrane potential will be reset to a fixed value after firing) and soft reset (membrane potential will subtract the threshold value after firing). 
$u_{\mathrm{reset}}$ denotes the reset membrane potential. 
Whenever the membrane potential $u^{t,l}$ exceeds the firing threshold $\theta^l$ at time $t$, the neuron will elicit an output spike $s^{t,l}$ to downstream spiking neurons, and the membrane potential $u^{t,l}$ will then reset to the reset potential $u_{\mathrm{reset}}$ (usually 0) or subtract the threshold value $\theta^l$. 

Direct training is an effective approach for training SNNs, which overcomes the non-differentiability of spiking activity through surrogate gradient functions, thereby enabling the application of the BPTT algorithm for SNNs. 
We obtain the update rule of weights as follows: 
\begin{equation}   
\label{deltaw}
        \mathbf{W}^l = \mathbf{W}^l - \eta\Delta\mathbf{W}^l,
\end{equation}  
\vspace{-2.5em}

\begin{equation}
\label{gradientw}
    \Delta \mathbf{W}^l = \frac{\partial \mathcal{L}}{\partial {\bm{{W}}^{l}}}=
       \sum_{t=1}^{T} \frac{\partial \mathcal{L}}{\partial {\bm{s}}^{{t}, {l}}} \frac{\partial {\bm{s}}^{{t}, {l}}}{\partial {\bm{u}}^{{t}, {l}}}
       \frac{\partial {\bm{u}}^{{t}, {l}}}{\partial {\bm{{W}}^{l}}}.
\end{equation}

Here $\mathcal{L}$ denotes the loss function, and $\eta$ represents the learning rate. 
According to Eq.(\ref{deltaw}), we could update network weights based on the gradient descent algorithm~\cite{sgd}. 
Eq.(\ref{gradientw}) represents the gradient of the loss with respect to the weights. 
Notably, we get the spike tensors $\bm{s}^{t, l}$ by non-differentiable Heaveside step function $H(\cdot)$, and researchers utilized differentiable surrogate gradient functions to replace the term $\frac{\partial \bm{s}^{t, l}}{\partial \bm{u}^{t, l}}$ to propagate the gradients during direct training. 
Two commonly used surrogate gradient functions can be described as follows: 

\begin{equation}
\label{rec}
    h_{\mathrm{rec}}(\bm{u^{t, l}}) = \frac{1}{\alpha}sign(|\bm{u^{t, l}}-\theta^l|<\frac{\alpha}{2}),
\end{equation}
\vspace{-2.0em}

\begin{equation}
\label{plg}
    h_{\mathrm{plg}}(\bm{u^{t, l}}) = \max(0, \alpha(1 - \alpha|\bm{u^{t, l}}-\theta^l|)).
\end{equation}

$h_{\mathrm{rec}}(\cdot)$ (Eq.(\ref{rec})) and $h_{\mathrm{plg}}(\cdot)$ (Eq.(\ref{plg})) respectively describe the derivative of the rectangular function \cite{STBP} and Piecewise Linear Grad (PLG) function \cite{plg}. 
$\alpha$ is a constant factor which determines the shape of function. 
We apply these two surrogate gradient functions in our work due to their widespread adoption and simplicity. 

\section{Methods}
\label{methods}

\subsection{Limited Information Representative Capacity of LIF Neuron Models}

\paragraph{Information loss caused by spiking activity}

\begin{figure}
\centering
\includegraphics[width=0.5\textwidth]{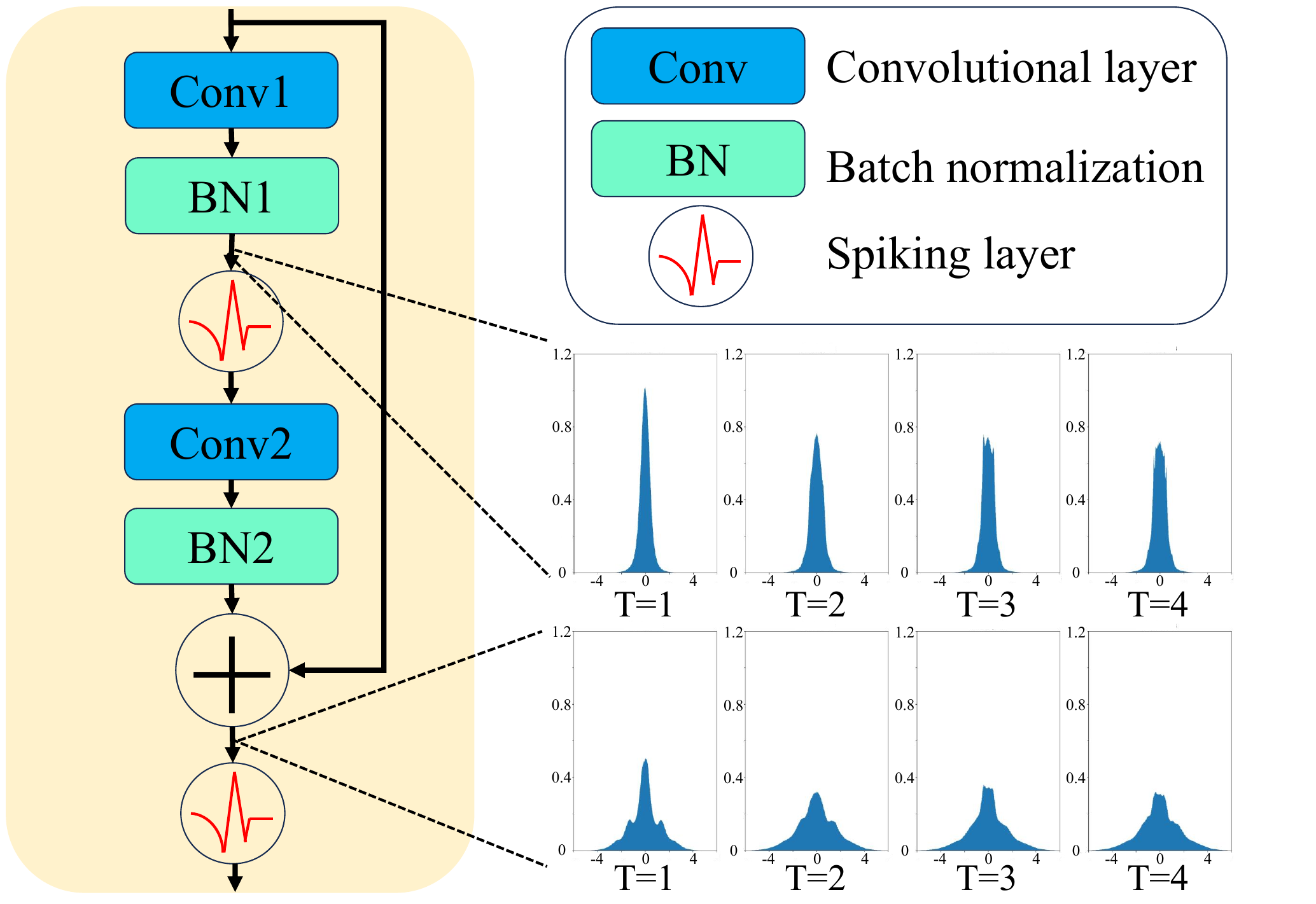}
\caption{The architecture of a residual block and the distribution of membrane potential before the two spiking layers.}
\label{residual}
\vspace{-1.0em} 
\end{figure}

The performance of SNNs still lags significantly behind that of ANNs, primarily due to substantial information loss in SNNs. 
The information loss in SNNs mainly originates from three sources. 
First, different from ANNs converting pre-activation values into continuous-valued activations via nonlinear functions, SNNs convert continuous-valued membrane potentials into binary spikes. 
Consequently, in contrast to ANNs, the binary spiking activity inherent to SNNs leads to substantial information loss. 
Second, the application of hard reset mechanism in direct training algorithms exacerbates the information loss in SNNs. 
When the membrane potential of LIF neuron model exceeds the threshold, LIF neuron will fire spikes to downstream neurons and the membrane potential will be reset to 0. 
Consequently, the extensive information encoded in  membrane potentials above threshold level (residual membrane potential)~\cite{rmp-loss, rmp-snn} is discarded. 
Third, in residual architectures, we found that the additive residual connection before spiking function increases the proportion of high membrane potentials. 
Fig.~\ref{residual} illustrates the difference in membrane potential distributions before and after the residual structure. 
We hypothesize that this disregards a larger proportion of the information retained in membrane potentials, which further intensifies information loss. 

\paragraph{The distribution of membrane potential}
In SNNs, the membrane potential serves as a fundamental state variable that bridges input stimuli and spike generation, playing a critical role in determining the network’s information representation and dynamic behavior.
The membrane potential dynamics of spiking neuron models are inherently complex. 
LIF neuron model will integrate the membrane potential from last time step and the normalized input current. 
We assume that the initial membrane potential is zero and we apply the tdBN method~\cite{tdBN}, and the pre-activations are normalized to $N(0, \theta^2)$.
Due to the sparsity of firing spikes, we neglect the reset mechanism of spiking neurons for simplicity.  
With the initial membrane potential $u^0 = 0$, Eq.~(\ref{lifupdate}) implies that $u^1 \sim N(0, \theta^2)$.
Due to the neglect of membrane potential reset mechanisms, we can get $u^2 \sim N(0, k_{\tau}^2 \theta^2 + \theta^2)$ and 
\begin{equation}
\label{mpd}
    u^t \sim N(0, (1 + \sum_{i=1}^{t-1}k_{\tau}^2)\theta^2).
\end{equation}
From Eq.(\ref{mpd}), we can observe the variations in the membrane potential distribution across different time steps. 

\paragraph{The impact of gradients in SNNs training}

As direct training methods update synaptic weights through gradient descent, the choice of surrogate gradient functions directly affects the backpropagation process. 
Many studies have explored how the graphical shapes and parameters of surrogate gradient functions influence the training of SNNs~\cite{STBP, adaptivesmooth}. 
In this study, our primary focus is to investigate the impact of gradient estimation on the performance of SNN models. 
We conduct probing experiments with two commonly used surrogate gradient functions: rectangular function~\cite{STBP} and PLG function~\cite{plg} under a spiking ResNet~\cite{tdBN} architecture. 
Tab.~\ref{SGresnet} represents the performance of SNN models with different parameters under a spiking ResNet architecture of surrogate functions. 
For spiking ResNet, the parameters of surrogate gradient functions have a notable impact on the performance of network. 
From the observations in Tab.~\ref{SGresnet}, we conclude that imprecise surrogate gradient approximations can significantly impair the performance of SNN models and even result in training failure. 
Therefore, it is essential to design more flexible, learnable surrogate gradient functions to optimize the direct training algorithm of SNNs. 


\begin{table}[htbp]\normalsize
    \centering
    \caption{The impact of gradient on the accuracy (\%) of ResNet-18 architecture on CIFAR-10 dataset}
    \begin{tabular}{ccccc}
    \toprule
    \hline
    SG functions & $\alpha$=0.5 & $\alpha$=1.0 & $\alpha$=2.0 & $\alpha$=5.0 \\
    \hline
    Rectangular & 93.99 & 94.84 & 93.28 &  36.92 \\
    PLG & 89.27 & 94.75 & 95.16 & 85.64 \\
    \hline
    \bottomrule
    \end{tabular}
    \vspace{-1.0em} 
    \label{SGresnet}
\end{table}
\vspace{-1.0em}

\begin{figure*}
\centering
\includegraphics[width=\textwidth]{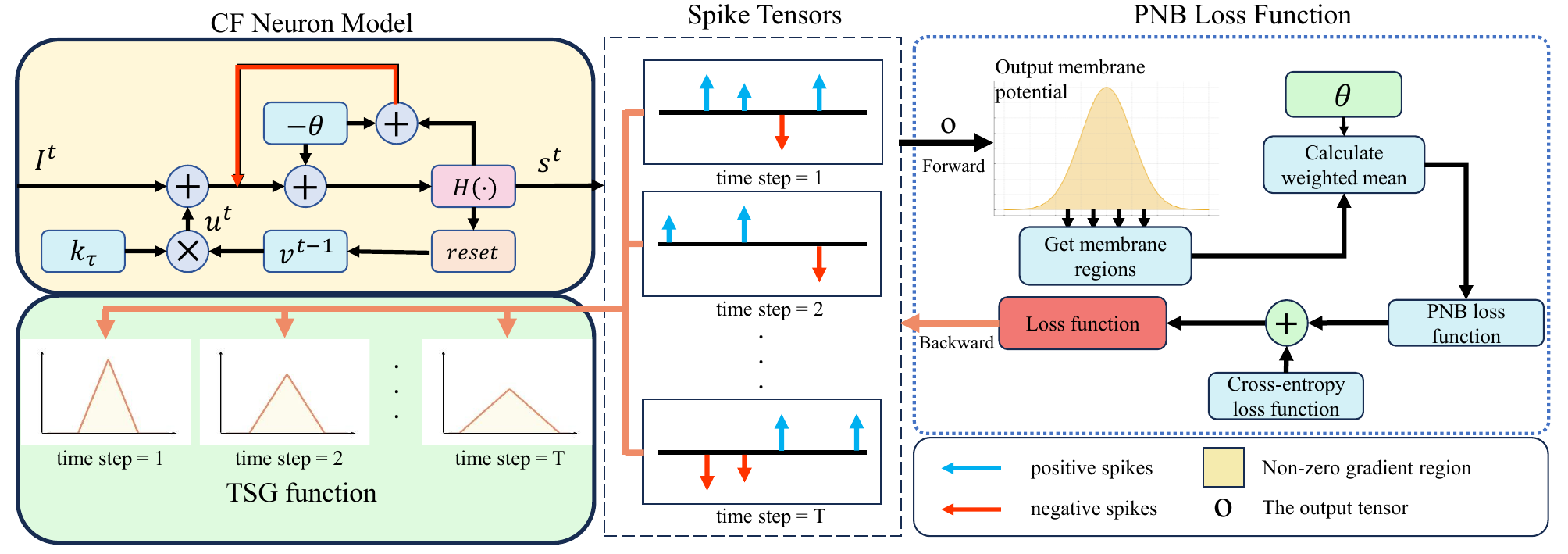}
\caption{The illustration of our direct training methods, including CF Neuron Model, TSG function and PNB loss function.}
\label{illustration}
\vspace{-1.0em} 
\end{figure*}

\subsection{Circulate-Firing (CF) Neuron Model}
\label{CF}
In this section, we introduce our circulate-firing (CF) neuron model which can enhance the information representation capacity of spiking neurons by effectively leveraging the rich information embedded in both the positive and negative regions of the membrane potential. 
We design circulate-firing mechanism to alleviate the information loss caused by the neglect of residual membrane potential. 
The expressions of CF neuron model are as follows




\begin{equation}
\label{cfreset}
    v_{i}^{t-1, l}=\begin{cases}
        \displaystyle u_{i}^{t-1, l}-s_{i}^{t-1, l}\theta_\mathrm{P}^l,\hspace{0.5em}  u_{i}^{t-1, l} > \theta_\mathrm{P}^{l}, \\
        \displaystyle u_{i}^{t-1, l}+s_{i}^{t-1, l}\theta_\mathrm{N}^l,\hspace{0.5em} u_{i}^{t-1, l} < \theta_\mathrm{N}^{l},\\
    \end{cases}
\end{equation}
\vspace{-2.0em}

\begin{equation}  
\label{cffiring}
    s_{i}^{t, l}=
    \begin{cases}
        \displaystyle \sum_{k_\mathrm{p}=1}^{K_\mathrm{P}} s_{i,k_\mathrm{p}}^{t,l},\hspace{0.5em}
        u_{i}^{t, l} > \theta_\mathrm{P}^{l}, \\[6pt]
        \displaystyle -\sum_{k_\mathrm{n}=1}^{K_\mathrm{N}}s_{i,k_\mathrm{n}}^{t,l},\hspace{0.5em}
        u_{i}^{t, l} < \theta_\mathrm{N}^{l}, \\[6pt]
        \displaystyle 0, \hspace{0.5em}\mathrm{otherwise},
    \end{cases}
\end{equation} 
\vspace{-1.0em}
    
\begin{equation}
\label{positive}
    s_{i,k_\mathrm{p}}^{t,l} = \mathbb{I} (u_i^{t,l} > k_\mathrm{p}\theta_\mathrm{P}^l), \hspace{0.5em}k_\mathrm{p}=1,\dots,K_\mathrm{P},
\end{equation}

\begin{equation}
\label{negative}
    s_{i,k_\mathrm{n}}^{t,l} = \mathbb{I} (u_i^{t,l} < k_\mathrm{n}\theta_\mathrm{N}^l), \hspace{0.5em}k_\mathrm{n}=1,\dots,K_\mathrm{N},
\end{equation}
\vspace{-2.0em}

\begin{equation}
\label{cfsg}
    h(u^{t, l}_i) = 
    \begin{cases}
        \displaystyle \alpha sign(|u^{t, l}_i - \frac{K_\mathrm{P} + 1}{2} \theta_\mathrm{P}^l| < \theta_\mathrm{P}^l\frac{K_\mathrm{P}}{2}), \hspace{0.5em} u^{t, l}_i \geq 0,  \\[4pt]
        \displaystyle \alpha sign(|u^{t, l} - \frac{K_\mathrm{N} + 1}{2} \theta_\mathrm{N}^l| < \theta_\mathrm{N}^l\frac{K_\mathrm{N}}{2}), \hspace{0.5em}\mathrm{otherwise.}
    \end{cases}
\end{equation}

Eq.(\ref{cfreset}) denotes the soft reset mechanism of membrane potential, and we apply soft reset to alleviate the information loss caused by the neglect of residual membrane potential. $\theta_\mathrm{P}^l$ and $\theta_\mathrm{N}^l$ respectively represent the positive and negative threshold in the $l-th$ layer. 
Eq.(\ref{cffiring})-Eq.(\ref{negative}) denotes the firing mechanism. 
$K_\mathrm{P}$ and $K_\mathrm{N}$ are the boundary numbers of circulate-firing mechanism in positive and negative membrane potential. 
$\mathbb{I}(\cdot)$ is an indicator function. $s_{i,k_\mathrm{p}}^{t,l}$ and $s_{i,k_\mathrm{n}}^{t,l}$ represent the $k_\mathrm{p}-$th positive spike and $k_\mathrm{n}-$th negative spike respectively. 
The number of firing spikes $s_{i}^{t, l}$ ranges from $-K_N$ to $K_P$. 
$h(\cdot)$ (Eq.(\ref{cfsg})) describes the surrogate gradient functions of CF neuron model, where $\alpha$ determines the gradient magnitudes of surrogate gradient functions. 

Our proposed CF spiking neuron model introduces a continuous membrane evolution mechanism, where the membrane potential is preserved and circulates across firing events rather than being fully reset. 
This spiking neuron model significantly reduces information loss, and thus, CF model could more effectively exploit accumulated membrane potential for subsequent spike generation. 
Such a mechanism could significantly enhance the information representation capacity of the network without increasing architectural complexity or parameter count. 
Fig.~\ref{illustration} provides an overview of our proposed methods. 

\subsection{Time-Step-Wise Learnable Surrogate Gradient Function}

Existing methods apply identical surrogate gradient functions across all time steps, ignoring the complex membrane potential dynamics at different time steps.
This neglect results in inaccurate gradient propagation and limits the performance of direct training approaches of SNNs. 
Moreover, spiking models employing fixed surrogate gradient functions tend to show inferior robustness against noise. 
To address these challenges, we propose a time-step-wise learnable surrogate gradient (TSG) function that varies across time steps as follows 


\begin{equation}
\label{positive tsg}
    h_{k_\mathrm{p}}(u^{t, l})=\max(0, \alpha^{t, l} (1 - \alpha^{t, l} |u^{t,l} - k_\mathrm{p}\theta_{\mathrm{P}}^l|), 
\end{equation}
\vspace{-2.0em}

\begin{equation}
\label{negative tsg}
    h_{k_\mathrm{n}}(u^{t, l})=\max(0, \alpha^{t, l} (1 - \alpha^{t, l} |u^{t,l} - k_\mathrm{n}\theta_{\mathrm{N}}^l|). 
\end{equation}

Eq.(\ref{positive tsg})-Eq.(\ref{negative tsg}) denote the TSG function, where $h_{k_\mathrm{p}}(u^{t, l})$ and $h_{k_\mathrm{n}}(u^{t, l})$ represent the surrogate gradient functions of $k_\mathrm{p}-$th positive spike and $k_\mathrm{n}-$th negative spike respectively. 
$\alpha^{t, l}$ denotes the parameter which determines the gradient of $l$-th layer in the $t$ time step. 

To enable stable and effective parameter optimization, a scale factor $s$, 
a bias term $b$, and a sigmoid activation $\sigma(\cdot)$ are introduced 
to parameterize $\alpha^{t,l}$ as
\begin{equation}
\alpha^{t,l} = s\,\sigma(x^{t,l}) + b,
\end{equation}
where $x^{t,l}$ denotes the learnable variable. 
This reparameterization constrains $\alpha^{t,l}$ within a controlled range, 
thereby preventing unstable updates during training while preserving 
sufficient flexibility for adaptation.

When the total number of time steps reduces to $T=1$, the proposed 
time-step-wise learnable surrogate gradient (TSG) naturally degenerates 
into a conventional learnable surrogate gradient. In this case, the 
formulation maintains enhanced gradient propagation capability compared 
with fixed (vanilla) surrogate gradient functions.

Conventional fixed surrogate gradients impose a predetermined response 
profile on spiking neurons, thereby limiting the adaptability of gradient 
flow. In contrast, the proposed TSG dynamically modulates its sensitivity 
to membrane potential variations across different time steps, enabling 
more flexible and effective temporal credit assignment.

\subsection{Positive-Negative Balanced Loss Function}

To enhance the expressive capacity of spiking neurons, we extend the conventional LIF model to CF spiking neuron model. 
Each spiking neuron is equipped with the ability to generate multiple spikes ($-K_\mathrm{N}$, $\cdots$, $-1$, $0$, $1$, $\cdots$, $K_\mathrm{P}$).

\begin{figure}
\begin{center}
\subfigure[Without PNB loss function]{
\includegraphics[width=1.6in,height=1in]{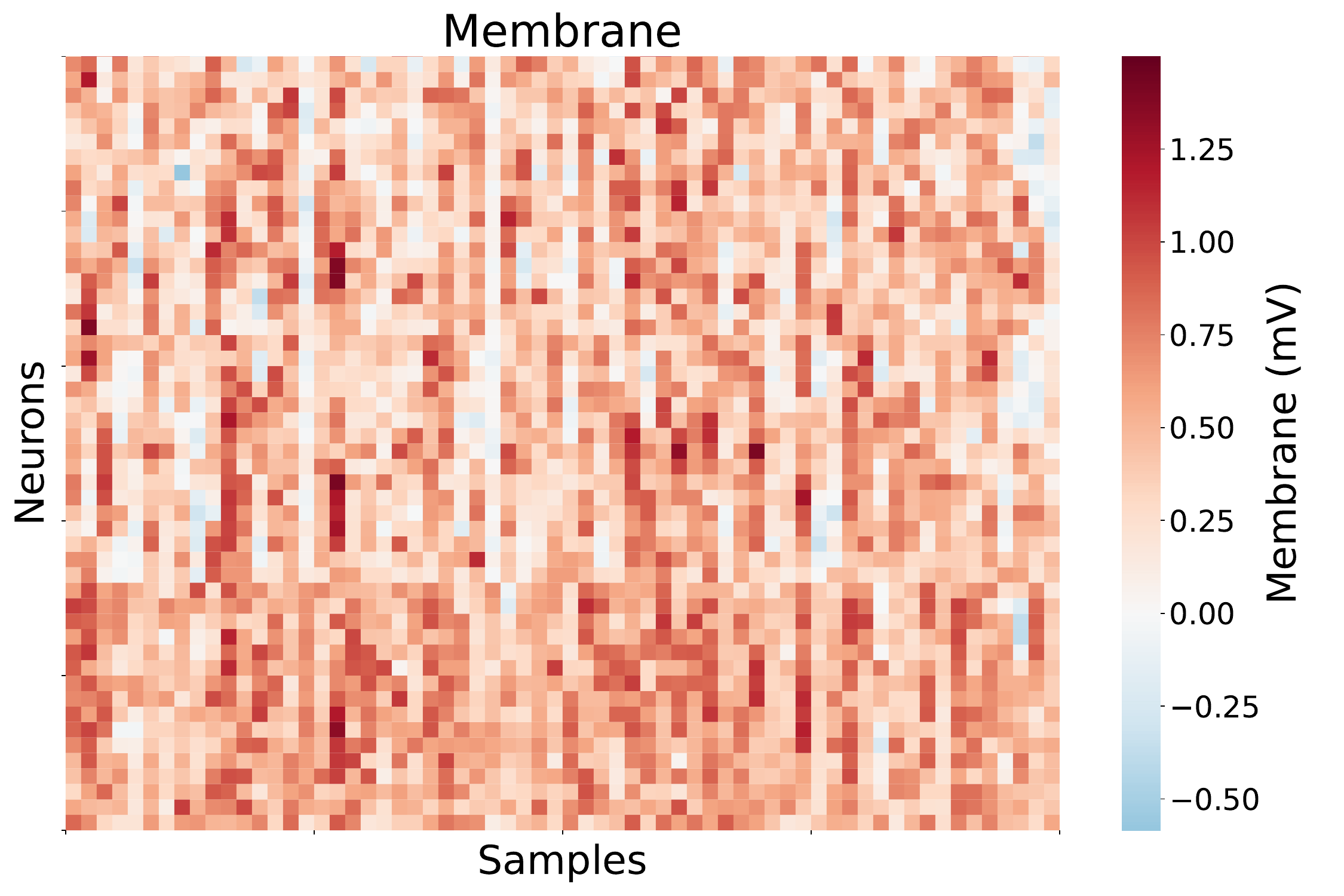}
}
\subfigure[With PNB loss function]{
\includegraphics[width=1.6in,height=1in]{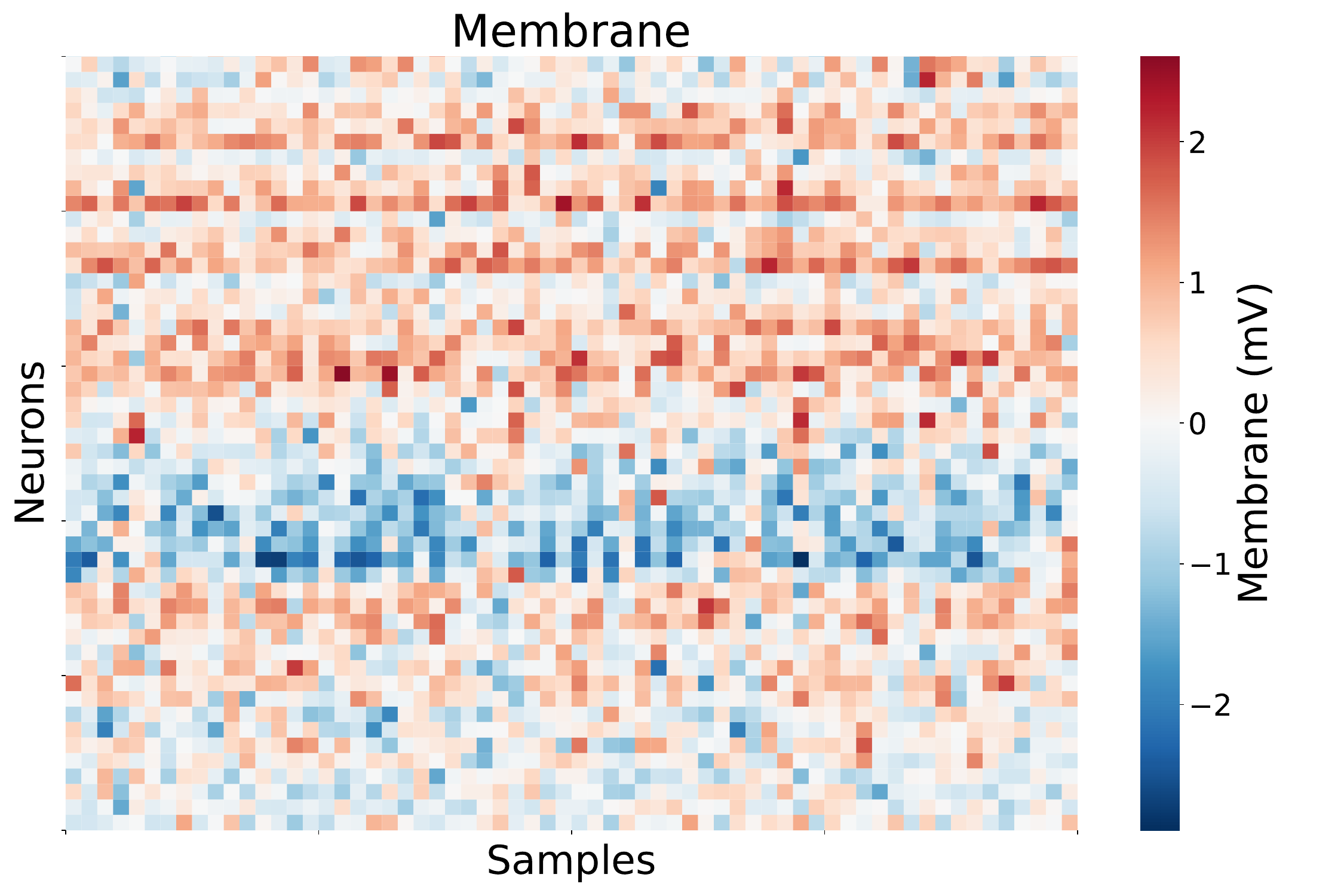}
}
\caption{The illustration of the magnitude of neuronal membrane potentials on the CIFAR-10 dataset using a ResNet-18 architecture.}
\label{mem}
\end{center}
\end{figure}

During training, we observe that the membrane potentials may drift towards excessively positive or negative regions, 
especially when multiple firing levels are introduced. 
To regularize the membrane dynamics, we propose a threshold-sensitive balancing loss that enforces symmetric statistics across all threshold intervals.
Specifically, we partition the positive membrane range into 
$K$ regions and construct the symmetric negative regions, 
\begin{equation}
    U^{+(k_\mathrm{p})} = \{u \mid {(k_\mathrm{p}-1)}\theta_{\mathrm{P}} < u \le k_\mathrm{p} \theta_{\mathrm{P}}\},
\end{equation}
\begin{equation}
    U^{-(k_\mathrm{n})} = \{u \mid -k_\mathrm{n} \theta_{\mathrm{N}} \le u < -(k_\mathrm{n}-1) \theta_{\mathrm{N}} \},
\end{equation}
where $U^{+(k_\mathrm{p})}$ denotes the $k_\mathrm{p}-$th positive region of membrane potential and $U^{-(k_\mathrm{n})}$ denotes the $k_\mathrm{n}-$th negative region of membrane potential. 
For each interval, we compute a weighted mean that emphasizes the membrane potentials close to firing:

\begin{equation}
    \mu^{+(k_\mathrm{p})} = 
    \frac{\sum_{u \in U^{+(k_\mathrm{p})}} u \cdot \exp\left(-\left|k_\mathrm{p} \theta_{\mathrm{P}} - u\right|\right)}
         {\sum_{u \in U^{+(k_\mathrm{p})}} \exp\left(-\left|k_\mathrm{p} \theta_{\mathrm{P}} - u\right|\right) + \varepsilon},
\end{equation}

\begin{equation}
    \mu^{-(k_\mathrm{n})} = 
    \frac{\sum_{u \in U^{-(k_\mathrm{n})}} u \cdot \exp\left(-\left|-k_\mathrm{n} \theta_{\mathrm{N}} - u\right|\right)}
         {\sum_{u \in U^{-(k_\mathrm{n})}} \exp\left(-\left|-k_\mathrm{n} \theta_{\mathrm{N}} - u\right|\right) + \varepsilon},
\end{equation}
where $\epsilon$ is a tiny constant for numerical stability. 
The multi-level membrane balancing loss is then defined as:
\begin{equation}
    \mathcal{L}_{\mathrm{pnb}} = 
    \frac{1}{K} \sum_{k=1}^{K}
    \left|\log\left(\frac{|\mu^{+(k_\mathrm{p})}|}{|\mu^{-(k_\mathrm{n})}| + \varepsilon} + \varepsilon\right)\right|,
\end{equation}
where $K=K_\mathrm{P}=K_\mathrm{N}$. 
Finally, the overall training objective is given by:
\begin{equation}
\label{loss}
    \mathcal{L} = \mathcal{L}_{\mathrm{CE}} + \lambda \cdot \mathcal{L}_{\mathrm{pnb}}.
\end{equation}
where $\mathcal{L}_{\mathrm{CE}}$ denotes the widely used cross entropy loss function. $\lambda$ is a tunable coefficient, and we set it as 0.25. 
This loss encourages symmetric membrane potential dynamics across all firing levels, thereby stabilizing the training of multi-threshold spiking neurons.
Fig.~\ref{mem} illustrates the magnitude comparison of membrane potential when the PNB loss function is applied versus when it is not. The horizontal axis represents different input samples and the vertical axis corresponds to individual neurons. Color intensity indicates the magnitude of the membrane potential. 

\subsection{Training Algorithm}

\begin{algorithm}
\caption{Training an SNN with CF spiking neural model, TSG function and PNB loss function.}
\label{alg}
\renewcommand{\algorithmicrequire}{\textbf{Input:}}
\renewcommand{\algorithmicensure}{\textbf{Output:}}
\begin{algorithmic}[1]
\REQUIRE
An SNN model to be trained with CF model, TSG function and PNB loss function; total number of training epochs N; total number of training iterations per epoch.
\ENSURE 
The trained SNN. 
\FOR{All $n=1, 2, \cdots, N-$th epoch}
\STATE \textbf{Feed-Forward:}
\STATE Get mini-batch training data $x(i)$ and class label $y(i)$;
\STATE Calculate the output of SNN model $O(i)$ and compute the classification loss $L=\mathcal{L}(O(i),y(i))$ by Eq.(\ref{alg output}) and Eq.(\ref{loss});
\STATE \textbf{Backward-Forward:}
\STATE Calculate the gradients of weights by Eq.(~\ref{gradientw});
\STATE Update the weights of model by Eq.(\ref{deltaw})
\ENDFOR
\end{algorithmic}
\end{algorithm}

In this part, we introduce the overall training algorithm based on the three aforementioned components. 
\subsubsection{Forward propagation}
In the forward propagation phase, the output of SNN model can be described as
\begin{equation}
\label{alg output}
    O = \frac{1}{T} \sum_{t=1}^{T} M  s ^{t, l_\mathrm{out}},
\end{equation}
where $O$ is the output membrane potential tensor, $T$ denotes the number of time steps, $M$ denotes the voting matrix connecting each voting neuron to a specific class, and $s ^{t, l_\mathrm{out}}$ denotes the output of the last spiking layer. 

As in previous methods, we adopt the cross-entropy loss function as follows,
\begin{equation}
\label{alg ce loss}
\mathcal{L}_{\mathrm{CE}} = - \sum_{c=1}^{C} y_c \log(p_c),
\end{equation}
where $C$ denotes the number of classes, and $y_c$ denotes the ground truth label of the $c-$th class, and $p_c$ is the predicted probability for the $c-$th class. 

\begin{table*}[htbp]\normalsize
\caption{Comparison results (\%) with previous state-of-the-art works on CIFAR-10/100 datasets. `ANN2SNN' denotes ANN-to-SNN conversion approach, `Direct Training' denotes direct training approach with surrogate gradient, `T' denotes `Time Step'.}
\label{comparison cifar}
\centering
\begin{tabular}{cccccc}
\toprule
\hline
Method & Type & Architecture & T & CIFAR-10 & CIFAR-100 \\
\cmidrule{1-6}
RMP-SNN \cite{rmp-snn} & ANN2SNN & VGG-13 & 2048 & 93.63 & 70.93 \\
\cmidrule{1-6}
\multirow{2}*{QCFS \cite{qcfs}} & \multirow{2}*{ANN2SNN} & VGG-16 & \multirow{2}*{32} & 95.66 & 77.01 \\
& & ResNet-20 &  & 93.25 & 69.82 \\
\cmidrule{1-6}
\multirow{2}*{SNN Calibration \cite{li2021calibration}} & \multirow{2}*{ANN2SNN} & ResNet-20 & \multirow{2}*{128} & 95.42 & 77.73 \\
& & VGG-16 &  & 95.65 & 77.40 \\
\cmidrule{1-6}
PLIF \cite{PLIF} & Direct Training & PLIFNet & 8 & 93.50 & $-$ \\
MLF \cite{MLF} & Direct Training & ResNet-20 & 4 & 94.25 & $-$ \\
\cmidrule{1-6}
\multirow{2}*{MPBN \cite{mpbn}} & \multirow{2}*{Direct Training} & \multirow{2}*{ResNet-19} & 1 & 96.06$\pm$0.10 & 78.71$\pm$0.10 \\
& & & 2 & 96.47$\pm$0.08 & 79.51$\pm$0.07 \\
\cmidrule{1-6}
\multirow{3}*{TAB \cite{TAB}} & \multirow{3}*{Direct Training} & \multirow{3}*{ResNet-19} & 2 & 94.73 & 76.31\\
& & & 4 & 94.76 & 76.81 \\
& & & 6 & 94.81 & 76.82 \\
\cmidrule{1-6}
\multirow{3}*{TET \cite{TET}} & \multirow{3}*{Direct Training} & \multirow{3}*{ResNet-19} & 2 & 94.16$\pm$0.03 & 72.87$\pm$0.10 \\
& & & 4 & 94.44$\pm$0.08 & 74.47$\pm$0.15 \\
& & & 6 & 94.50$\pm$0.07 & 74.72$\pm$0.28 \\
\cmidrule{1-6}
\multirow{3}*{RMP-Loss \cite{rmp-loss}} & \multirow{3}*{Direct Training} & \multirow{3}*{ResNet-19} & 2 & 95.31$\pm$0.07 & 74.66$\pm$0.12 \\
& & & 4 & 95.51$\pm$0.08 & 78.28$\pm$0.10 \\
& & & 6 & 96.10$\pm$0.08 & 78.98$\pm$0.08 \\
\cmidrule{1-6}
\multirow{4}*{IM-Loss \cite{imloss}} & \multirow{4}*{Direct Training} & \multirow{3}*{ResNet-19} & 2 & 93.85$\pm$0.10 & $-$ \\
& & & 4 & 95.40$\pm$0.08 & $-$ \\
& & & 6 & 95.49$\pm$0.05 & $-$ \\
& & VGG-16 & 5 & 93.85$\pm$0.11 & 70.18$\pm$0.09 \\
\cmidrule{1-6}
\multirow{2}*{ASGL \cite{adaptivesmooth}} & \multirow{2}*{Direct Training} & \multirow{2}*{ResNet-18} & 2 & 95.27$\pm$0.06 & 76.59$\pm$0.05 \\
& & & 4 & 95.35$\pm$0.25 & 77.74$\pm$0.07 \\
\cmidrule{1-6}
\multirow{2}*{LM-H \cite{lm-h}} & \multirow{2}*{Direct Training} & ResNet-18 & 4 & 95.32 & 78.58 \\
& & ResNet-19 & 4 & 96.36 & 81.65 \\
\cmidrule{1-6}
\multirow{2}*{LM-HT~\cite{lm-ht}} & \multirow{2}*{Direct Training} & ResNet-18 & 2 & 96.25 & 79.33 \\
& & ResNet-19 & 2 & 96.89 & 81.76 \\
\cmidrule{1-6}
\multirow{3}*{\textbf{Ours}} & \multirow{3}*{\textbf{Direct Training}} & \multirow{3}*{\textbf{ResNet-18}} & \textbf{1} & \textbf{96.84$\pm$0.12} & \textbf{79.56$\pm$0.12} \\
&  &  & \textbf{2} & \textbf{97.19$\pm$0.08} & \textbf{80.21$\pm$0.11} \\
&  &  & \textbf{4} & \textbf{97.31$\pm$0.11} & \textbf{80.89$\pm$0.18} \\
\hline
\bottomrule
\end{tabular}
\vspace{-1.0em} 
\end{table*}

To balance the positive and negative membrane potential, we introduce PNB loss function to the overall loss function.  

\subsubsection{Backward propagation}

In the backward propagation, we apply the TSG function to address the non-differentiability of spiking activity. 
We apply direct training approach to train our SNN models based on Eq.\ref{deltaw}-Eq.\ref{gradientw}.
Algorithm~\ref{alg} represents the overall training process of our method. 

\section{Experimental Results}
\label{experimental results}

In this section, we conduct extensive experiments on multiple static datasets, neuromorphic datasets, text classification datasets and tactile datasets to demonstrate the effectiveness of our proposed methods. 
We compare the results of our algorithm with previous state-of-the-art (SoTA) methods and ANNs models. 
Ablation studies were conducted to assess the effectiveness of our methods, and noise robustness evaluations demonstrate that our algorithm possesses superior noise resistance capability.

\subsection{Experimental Setup}

We evaluate our proposed method with CNN architecture (ResNet \cite{he2016resnet}, VGG \cite{simonyan2014vgg}) and Transformer architecture (Spikformer \cite{zhou2022spikformer}) based on various static, neuromorphic, text classification datasets and tactile datasets, including CIFAR-10/100, ImageNet-200, DVS-CIFAR10, DVS Gesture, SST-2/5, Waimai, EvTouch-Objects and EvTouch-Containers datasets. 

\subsubsection{Datasets}
The CIFAR-10 and CIFAR-100 datasets \cite{krizhevsky2009cifar} are widely used benchmarks for image classification. 
Both datasets consist of 60,000 color images with a spatial resolution of 32 $\times$ 32, split into 50,000 training samples and 10,000 test samples. 
CIFAR-10 contains 10 mutually exclusive classes. 
CIFAR-100 includes 100 fine-grained categories with 600 images per class, making it a more challenging benchmark due to its higher inter-class similarity and lower sample size per category. 
ImageNet-200~\cite{deng2009imagenet} dataset is a widely used subset of the original ImageNet-1K benchmark. 
It contains 200 object categories selected from the full 1,000 classes. 
The dataset consists of approximately 128,000 training images and 10,000 validation images, all in RGB format with varying resolutions. 
Compared to ImageNet-1K, ImageNet-200 retains high visual diversity while offering reduced computational cost, making it a suitable benchmark for evaluating scalable yet efficient learning algorithms.
The DVS-CIFAR10~\cite{li2017dvscifar10} dataset is the neuromorphic version of the CIFAR-10 dataset, which includes 10 000 images. 
We split the dataset into 9000 images for training and 1000 images for testing. 
The DVS Gesture~\cite{gesture} dataset is recorded by a DVS128 camera, which contains 11 kinds of hand gestures from 29 subjects under 3 kinds of illumination conditions. 
The SST-2/5~\cite{sst} datasets are widely used benchmarks for sentence-level sentiment classification. 
SST-2 is the binary classification variant, in which only positive and negative samples are retained, with neutral instances removed. 
It is commonly used to evaluate polarity detection in short, user-generated texts.
SST-5 preserves the original fine-grained five-class annotation scheme, including very positive, positive, neutral, negative, and very negative. 
Compared to SST-2, SST-5 poses a more challenging task that requires models to capture subtle sentiment differences. 
Waimai dataset consists of 12, 000 Chinese user reviews collected by a food delivery platform for binary sentiment classification (positive and negative).
EvTouch-Objects dataset comprises tactile data from 36 object classes. 
EvTouch-Containers dataset includes tactile data for four containers: a coffee can, a plastic soda bottle, a soymilk carton, and a metal tuna can. 
Following previous work~\cite{gu2020tactilesgnet}, we use 80\% of the samples for training and the remaining 20\% for testing.

\begin{table}[htbp]\normalsize
\caption{Comparison with ANN models and Transformer-based models on CIFAR-10/100 datasets. `Param' denotes `Parameter (M)', `T' denotes `Time Step', `CIFAR-10/100' denotes `CIFAR-10/100 Top-1 Accuracy (\%)'.}
\label{comparison ANN Transformer cifar}
\centering
\begin{tabular}{cccc}
\toprule
\hline
Method & Param & T & CIFAR-10/100 \\
\cmidrule{1-4}
Spikformer \cite{zhou2022spikformer} & 9.32    & 4  &  95.51 / 78.21 \\
QKFormer \cite{zhou2024qkformer} & 6.74    & 4  & 96.18 / 81.15 \\
Spikformer + SEMM \cite{SEMM} &  $-$  & 4  & 96.16 / 80.24 \\
\cmidrule{1-4}
ResNet-19 (ANN) \cite{TET} & 12.63  &1 & 94.97 / 75.35 \\
Transformer (ANN)\cite{zhou2024qkformer} &  9.32  &1 & 96.73 / 81.02 \\
ResNet-18 (ANN) & 11.17 & 1 & 96.70 / 79.84 \\
\cmidrule{1-4}
\textbf{Spikformer (ours)} & \textbf{9.33} & \textbf{4} & \textbf{97.06 / 81.09} \\
\textbf{ResNet-18 (ours)} & \textbf{11.17} & \textbf{4} & \textbf{97.31 / 80.89} \\
\hline
\bottomrule
\end{tabular}
\vspace{-1.0em} 
\end{table}

\begin{table*}[!t]\normalsize
\caption{Comparison results (\%) with previous state-of-the-art works on ImageNet-200 dataset. `ANN2SNN' denotes ANN-to-SNN conversion approach, `Direct Training' denotes direct training approach with surrogate gradient, `Hybrid' denotes hybrid training approach, `T' denotes `Time Step'.}
\label{comparison imagenet200}
\centering
\begin{tabular}{ccccc}
\toprule
\hline
Method & Type & Architecture & T & ImageNet-200 \\
\cmidrule{1-5}
QCFS \cite{qcfs} & ANN2SNN & VGG-16 & 32 & 53.54 \\
SNN calibration \cite{li2021calibration} & ANN2SNN & VGG-16 & 32 & 53.96 \\
Online-LTL \cite{onoffline} & Hybrid & VGG-13 & 16 & 54.82 \\
Offline-LTL \cite{onoffline} & Hybrid & VGG-13 & 16 & 55.37 \\
LM-H \cite{lm-h} & Direct Training & VGG-13 & 4 & 59.93 \\
\multirow{2}*{ASGL \cite{adaptivesmooth}} & \multirow{2}*{Direct Training} & \multirow{2}*{VGG-13} & 4 & 56.57 \\
& & & 8 & 56.81 \\
\multirow{2}*{LM-HT \cite{lm-ht}} & \multirow{2}*{Direct Training} & \multirow{2}*{VGG-13} & 2 & 61.09 \\
& & & 4 & 61.75 \\
\cmidrule{1-5}
\multirow{3}*{\textbf{Ours}} & \multirow{3}*{\textbf{Direct Training}} & \multirow{3}*{\textbf{VGG-11}} & \textbf{1} & \textbf{61.07$\pm$0.06} \\
&  &  & \textbf{2} & \textbf{61.76$\pm$0.09} \\
&  &  & \textbf{4} & \textbf{62.14$\pm$0.08} \\
\hline
\bottomrule
\end{tabular}
\vspace{-1.0em} 
\end{table*}

\subsubsection{Setup of experimental parameters}

For static datasets such as CIFAR-10, CIFAR-100, and ImageNet-200, models are trained from scratch following the data augmented methods as previous works~\cite{TAB}.
For neuromorphic datasets such as DVS-CIFAR10, we applied the data augmented methods in~\cite{NDA}. 
For text classification datasets such as SST-2/5 and Waimai, we employ word embedding and spike generation methods similar to those used in previous works~\cite{snntext}. 
All experiments are conducted using PyTorch on NVIDIA RTX4090 GPUs. 
Network models are trained using the SGD optimizer with an initial learning rate of 0.025 and a batch size of 64 for 400 epochs. 
The momentum of SGD is 0.9, and the weight decay for static datasets is 1e-4, and the weight decay for neuromorphic datasets is 5e-4. 
We apply the CosineAnnealingLR learning rate scheduler to adjust the learning rate during training. 
The decay factor of spiking neuron models is set to 0.25, and $\theta_P$ is set to 1.0, and $\theta_N$ is set to -1.0, and $K_\mathrm{P}$ and $K_\mathrm{N}$ are set to 2. 
For CIFAR-10/100 datasets, we apply ResNet architecture, for ImageNet-200 dataset and neuromorphic datasets, we apply VGG architecture, for text classification datasets, we apply TextCNN~\cite{textcnn} architecture, for tactile datasets, we apply 4-layer CNN architecture. All results are reported as the average over three independent runs to ensure statistical reliability. 
\subsection{Comparison with Previous SoTA Works}

We performed extensive experiments on static datasets (CIFAR-10/100~\cite{krizhevsky2009cifar} and ImageNet-200~\cite{deng2009imagenet}) and neuromorphic datasets (DVS-CIFAR10~\cite{li2017dvscifar10} and DVS Gesture~\cite{gesture}) with architectures that are consistent with previous works, including VGG~\cite{simonyan2014vgg}, ResNet~\cite{he2016resnet} and Spikformer~\cite{zhou2022spikformer}. 
To comprehensively evaluate the effectiveness of the proposed method, we compare our approach with a wide range of state-of-the-art (SoTA) spiking neural network training algorithms, including ANN-to-SNN conversion methods and directly trained SNNs with surrogate gradients.

\paragraph{CIFAR-10/100}
Tab.~\ref{comparison cifar} reports the classification accuracy of different methods on CIFAR-10 and CIFAR-100 under various simulation time steps. Our method consistently outperforms existing SoTA approaches across all evaluated time steps. 
Specifically, under the extremely low-latency setting of one time step, our method achieves 96.84\% on CIFAR-10 and 79.56\% on CIFAR-100, surpassing previous direct-training methods by a significant margin. As the number of time steps increases, our advantage becomes more pronounced. With 4 time steps, our method reaches 97.31\% / 80.89\%. 

To further assess the generalization ability of our approach, we evaluate it on Transformer-based architectures and compare it with both ANN and SNN Transformers, as summarized in Tab.~\ref{comparison ANN Transformer cifar}.
When applied to Spikformer, our method improves the accuracy to 97.06\% on CIFAR-10 and 81.09\% on CIFAR-100, outperforming existing SNN Transformer baselines such as Spikformer~\cite{zhou2022spikformer} and QKFormer~\cite{zhou2024qkformer}. Notably, our spiking models achieve comparable or even superior accuracy to ANN counterparts with identical architectures, highlighting the effectiveness of the proposed neuron model and training strategy in closing the performance gap between SNNs and ANNs. 
These results demonstrate that our method not only improves accuracy but also significantly enhances the accuracy–latency efficiency and generalizability of SNNs.


\paragraph{ImageNet-200}
We further evaluate our method on the more challenging ImageNet-200 dataset to assess its scalability. As shown in Tab.~\ref{comparison imagenet200}, our approach achieves 61.07\% accuracy with only one time step using a VGG-11 architecture, already surpassing several prior methods that require deeper networks and significantly more time steps.

With 4 time steps, our method reaches 62.14\%, outperforming recent SoTA direct-training approaches such as ASGL~\cite{adaptivesmooth} and LM-HT~\cite{lm-ht}. These results indicate that the proposed method scales effectively to larger datasets while maintaining high efficiency under low-latency settings.

\begin{table}[htbp]\normalsize
\caption{Comparison results (\%) with previous state-of-the-art works on DVS-CIFAR10 dataset. `T' denotes `Time Steps'. `Acc' denotes `Top-1 Accuracy (\%)'.\label{comparison dvscifar10}}
\centering
\begin{tabular}{cccc}
\toprule
\hline
Method & Architecture & T & Acc \\
\cmidrule{1-4}
PLIF \cite{PLIF} & PLIF Net & 20 & 74.80 \\
GLIF \cite{yao2022glif} & Wide 7B Net & 16 & 78.10 \\
TET \cite{TET} & VGG & 10 & 83.17 \\
TEBN \cite{tebn} & VGG & 10 & 84.90 \\
PSN \cite{psn} & VGG & 10 & 85.90 \\
NDA \cite{NDA} & VGG & 10 & 81.70 \\
EventMix \cite{shen2023eventmix} & ResNet-18 & 10 & 81.45 \\
EventRPG \cite{sun2024eventrpg} & VGG & 10 & 84.96 \\
TCJA \cite{zhu2024tcja} & TCJA-SNN & 10 & 83.30 \\
\cmidrule{1-4}
\textbf{Ours} & \textbf{VGG} & \textbf{10} & \textbf{87.00$\pm$0.14} \\
\hline
\bottomrule
\end{tabular}
\end{table}

\begin{table}[htbp]\normalsize
\caption{Comparison with Transformer-based models on DVS-CIFAR10 dataset. `Param' denotes `Parameter (M)', `T' denotes `Time Steps', `Acc' denotes `Top-1 Accuracy (\%)'}
\label{comparison ANN Transformer dvscifar10}
\centering
\begin{tabular}{cccc}
\toprule
\hline
Method & Param & T & Acc \\
\cmidrule{1-4}
Spikformer \cite{zhou2022spikformer} & 2.57 & 16  &  80.90 \\
QKFormer \cite{zhou2024qkformer} & 1.50    & 4  & 84.00 \\
SpikFormer + SEMM \cite{SEMM} &  $-$  & 16  & 82.10 \\
\cmidrule{1-4}
\textbf{Spikformer (ours)} & \textbf{2.57} & \textbf{10} & \textbf{85.07$\pm$0.17} \\
\textbf{VGG (ours)} & \textbf{9.30} & \textbf{10} & \textbf{87.00$\pm$0.14} \\
\hline
\bottomrule
\end{tabular}
\end{table}

\begin{table}[htbp]\normalsize
\caption{Comparison results (\%) with previous state-of-the-art works on DVS Gesture dataset. `T' denotes `Time Steps'. `Acc' denotes `Top-1 Accuracy (\%)'\label{comparison dvsgesture}}
\centering
\begin{tabular}{cccc}
\toprule
\hline
Method & Architecture & T & Acc \\
\cmidrule{1-4}
PLIF \cite{PLIF} & PLIF Net & 20 & 97.57 \\
EventMix \cite{shen2023eventmix} & ResNet-18 & 10 & 96.75 \\
TCJA \cite{zhu2024tcja} & TCJA-SNN & 10 & 98.20 \\
\cmidrule{1-4}
\textbf{Ours} & \textbf{VGG} & \textbf{10} & \textbf{98.33$\pm$0.21} \\
\hline
\bottomrule
\end{tabular}
\vspace{-1.0em}
\end{table}

\paragraph{Neuromorphic datasets}
We also evaluate the performance of our methods on the neuromorphic dataset. 
As shown in Tab.~\ref{comparison dvscifar10}, we achieve the accuracy of 87.00\% with 10 time steps, surpassing previous state-of-the-art approaches (PSN~\cite{psn}) by 1.10\%. 
To demonstrate the generalization ability of our methods, we conduct experiments under Transformer-based architecture. 
As shown in Tab.~\ref{comparison ANN Transformer dvscifar10}, with equivalent parameter size (2.57M) and fewer time steps, we improve the accuracy of 4.20\% compared to baseline~\cite{zhou2022spikformer}. 
Consequently, our algorithm achieves both high performance and broad compatibility with convolutional neural networks (CNNs) and Transformers architectures. 

As shown in Tab.~\ref{comparison dvsgesture}, on the DVS Gesture dataset, our method reaches 98.33\%, achieving the best reported performance among direct-training SNNs.

\paragraph{Text classification}
To validate cross-modal generalization, we further evaluate our method on text sentiment classification (SST-2/5 and Waimai). 
As shown in Tab.~\ref{comparison nlp}, our SNN model outperforms the previous converted-and-finetuned SNN models. This improvement indicates that our method can learn more discriminative representations on text datasets. Moreover, the consistent gains across multiple datasets further demonstrate its strong cross-modal generalization ability and effectiveness.

\begin{table*}[htbp]\normalsize
\caption{Comparison with previous state-of-the-art works on natural language sentiment classification datasets. `ANN' denotes ANN training approach, `ANN2SNN' denotes ANN-to-SNN conversion approach, `Direct Training' denotes direct training approach with surrogate gradient, `Hybrid' denotes hybrid training approach, `T' denotes `Time Steps'.}
\label{comparison nlp}
\centering
\begin{tabular}{cccccc}
\toprule
\hline
Method & Type & T & SST-2 & SST-5 & Waimai \\
\cmidrule{1-6}
Original TextCNN & ANN & $-$ & 83.25$\pm$0.16 & 45.48$\pm$0.16 & 88.49$\pm$0.16 \\
Tailored TextCNN & ANN & $-$ & 83.03$\pm$0.21 & 43.48$\pm$0.13 & 88.21$\pm$0.15 \\
Directly-trained SNN & Direct Training & 50 & 75.73$\pm$0.91 & 23.08$\pm$0.56 & 66.42$\pm$0.39 \\
Conv SNN & ANN2SNN & 50 & 80.07$\pm$0.78 & 41.40$\pm$0.73 & 86.43$\pm$0.43 \\
Conv SNN + FT & Hybrid & 50 & 80.91$\pm$0.34 & 41.63$\pm$0.44 & 86.66$\pm$0.17 \\
\cmidrule{1-6}
\textbf{Ours} & \textbf{Direct Training} & \textbf{50} & \textbf{81.31$\pm$0.29} & \textbf{41.90$\pm$0.26} & \textbf{86.89$\pm$0.32} \\
\hline
\bottomrule
\end{tabular}
\vspace{-1.0em} 
\end{table*}

\paragraph{Tactile classification}
We also conduct experiments on tactile datasets (EvTouch-Objects and Evtouch-Containers) to demonstrate the effectiveness of our proposed method. 
Tab.~\ref{comparison tactile} shows that our method consistently outperforms existing approaches on both datasets. This improvement indicates that our model possesses stronger representation capability and can more effectively capture the spatiotemporal characteristics of tactile datasets. Furthermore, the superior performance demonstrates its effectiveness across different modalities.  

These results reflect that the proposed neuron model and learning strategy are not limited to visual tasks, but generalize well to other modalities. 

Overall, our proposed method achieves consistent and substantial improvements across static images, neuromorphic vision, natural language, and tactile tasks. Our performance gains stem from enhanced neuron-level information representation and more accurate temporal gradient propagation, enabling superior performance under low latency and diverse application scenarios.

\begin{table}[htbp]\normalsize
\caption{Comparison results (\%) with previous methods on EvTouch-Objects and EvTouch-Containers datasets. `Objects' denotes `EvTouch-Objects', and `Containers' denotes `EvTouch-Containers'.\label{comparison tactile}}
\centering
\begin{tabular}{ccc}
\toprule
\hline
Method & Objects & Containers\\
\cmidrule{1-3}
TactileGCN~\cite{garcia2019tactilegcn} & 66.70 & 51.30 \\
SLAYER~\cite{shrestha2018slayer} & 81.40 & 65.00 \\
Grid-based CNN~\cite{gu2020tactilesgnet} & 88.40 & 60.17 \\
MLP~\cite{gu2020tactilesgnet} & 85.97 & 58.83 \\
TactileSGNet~\cite{gu2020tactilesgnet} & 89.44 & 64.17 \\
\hline
\textbf{Ours} & \textbf{90.28} & \textbf{70.00} \\
\hline
\bottomrule
\end{tabular}
\vspace{-1.0em} 
\end{table}

\subsection{Ablation Study}

In this section, we conduct a comprehensive ablation study to systematically analyze the contribution and underlying mechanism of each proposed component, including the circulate-firing (CF) neuron model, the time-step-wise surrogate gradient (TSG) function, and the positive–negative balanced (PNB) loss. All ablation experiments are conducted under identical training settings to ensure a fair comparison.

\subsubsection{The contribution of each component}

\begin{figure}[htbp]\normalsize
  \centering
  \includegraphics[width=0.48\textwidth]{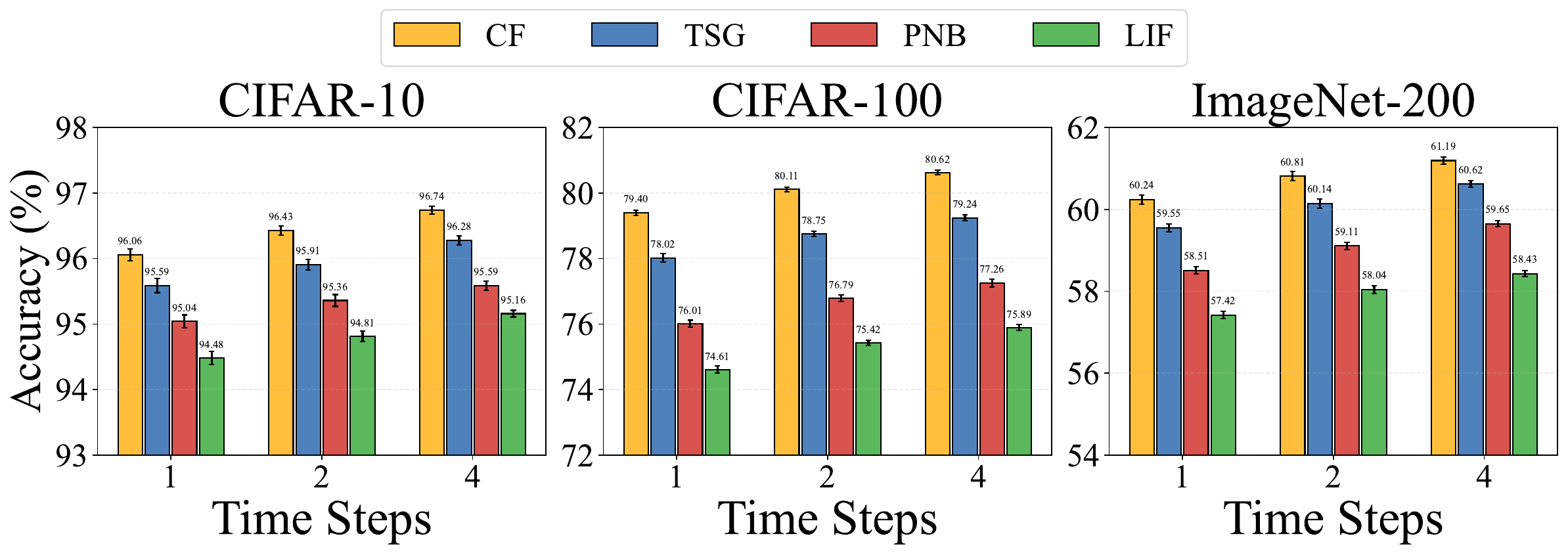}
  \caption{Accuracy comparison of the baseline LIF-based model 
and the corresponding variants formed by individually adding 
each proposed component, including the CF neuron, 
the time-step-wise surrogate gradient (TSG), and the PNB loss.}
  \label{ablation}
  \vspace{-1.0em} 
\end{figure}

\begin{figure}[htbp]\normalsize
  \centering
  \includegraphics[width=0.48\textwidth]{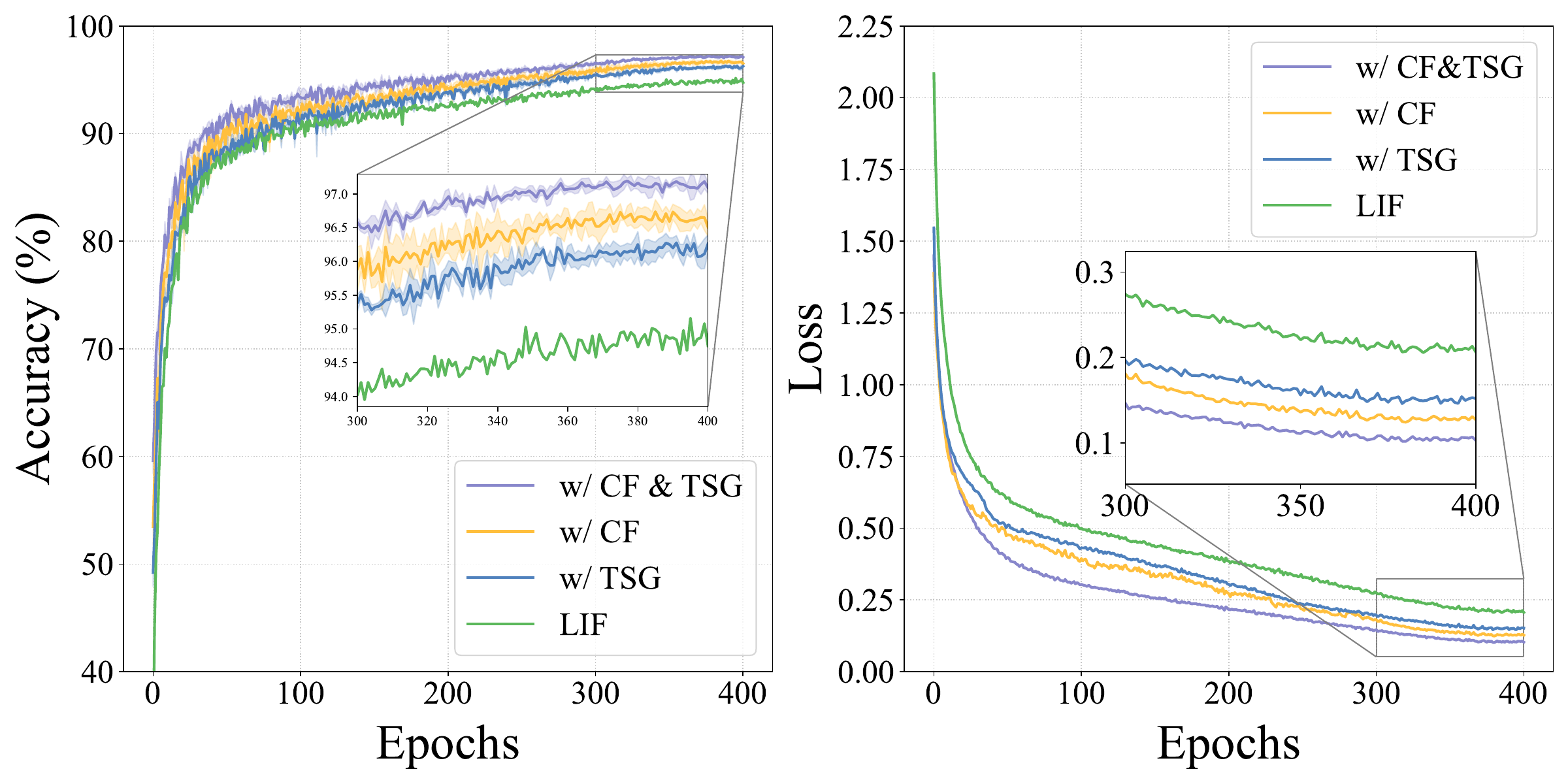}
  \caption{The test accuracy and loss curves on CIFAR-10 with ResNet-18 
for the baseline model and its three variants obtained by 
individually incorporating each proposed component.}
  \label{ablation_loss}
  \vspace{-1.0em} 
\end{figure}

We first evaluate the independent contribution of each proposed component 
by introducing them separately into the baseline model. 
Fig.~\ref{ablation} summarizes the classification accuracy on CIFAR-10, 
CIFAR-100, and ImageNet-200 under different simulation time steps.

Starting from the baseline equipped with the vanilla LIF neuron, 
we construct three variants by independently incorporating 
(i) the CF neuron, 
(ii) the time-step-wise surrogate gradient (TSG), and 
(iii) the PNB loss, respectively. 

Replacing the vanilla LIF neuron with the CF neuron consistently 
improves performance across all datasets and time steps, 
indicating that enhancing the information representation capability 
of spiking neurons effectively strengthens the discriminative capacity 
of SNNs. 

When the TSG function is applied to the baseline while keeping the 
original neuron model and loss unchanged, accuracy is also improved, 
demonstrating that adaptive temporal gradient modulation facilitates 
more effective error propagation. 

Similarly, incorporating the PNB loss alone into the baseline model 
brings measurable performance gains and improves training stability, 
suggesting that it provides beneficial optimization guidance.

Fig.~\ref{ablation_loss} further presents the accuracy and loss curves 
on CIFAR-10 with ResNet-18, comparing the baseline model and three 
variants constructed by individually introducing each proposed 
component (CF neuron, TSG, or PNB) into the baseline framework.

Overall, each component, when individually incorporated into the baseline, 
consistently improves performance, demonstrating its standalone effectiveness. 
Moreover, combining multiple components leads to further performance gains, 
with the complete model achieving the best results. These observations 
indicate that the three modules provide complementary benefits and 
contribute cumulatively to the final improvement.


\subsubsection{Information representation capacity of CF neurons}

\begin{figure}[htbp]
\centering
\includegraphics[width=0.48\textwidth]{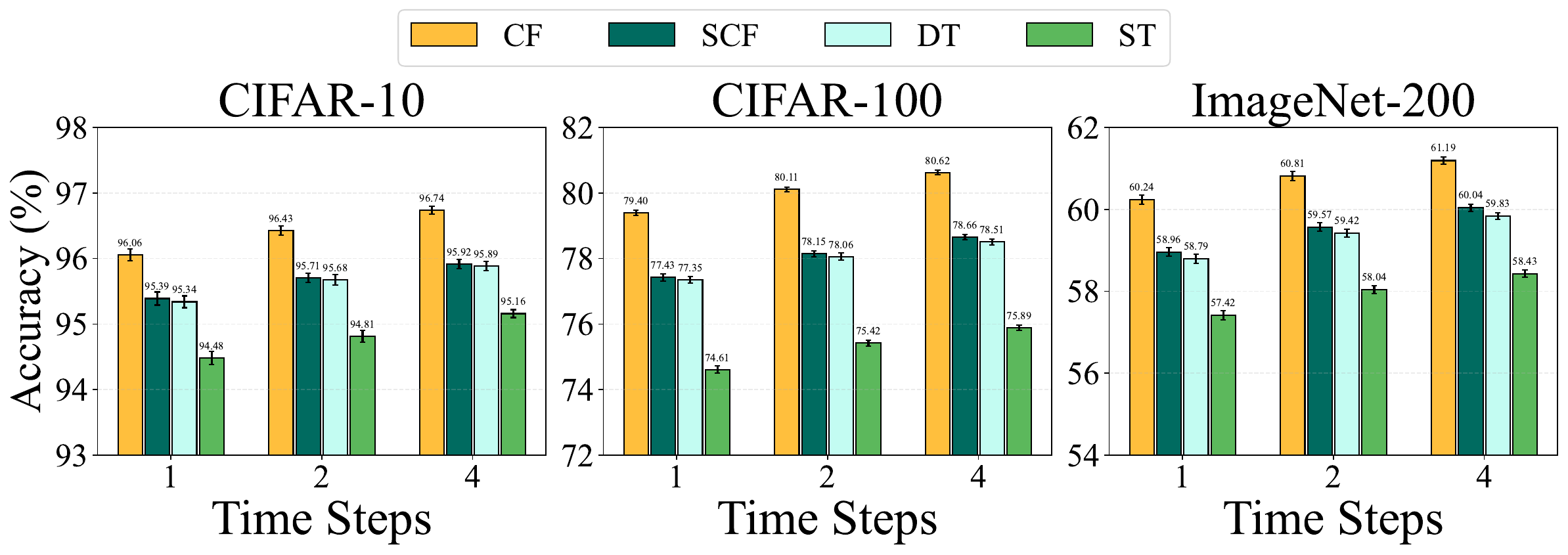}
\caption{Accuracy comparison of four spiking neuron models (ST, DT, SCF, and CF) on CIFAR-10, CIFAR-100, and ImageNet-200 datasets under different time steps.}
\label{ablation_spike}
\vspace{-1.0em} 
\end{figure}




To further investigate the effectiveness of our CF neuron model, we compare it with three alternative spike generation mechanisms: single-threshold (ST), double-threshold (DT), and single circulate-firing (SCF). The comparison results are shown in Fig.~\ref{ablation_spike}.

Compared with the conventional single-threshold LIF neuron, CF neurons consistently achieve higher accuracy across all datasets and time steps. Notably, while DT and SCF partially improve performance by exploiting either negative membrane potentials or multi-level firing, CF neurons simultaneously leverage both positive and negative membrane potential regions with multiple firing levels, leading to the most expressive spike representation.

These results demonstrate that CF neurons significantly enhance the effective information utilization of membrane potentials, thereby reducing the information loss caused by binary spike conversion.

\begin{figure}[htbp]
\centering
\includegraphics[width=0.48\textwidth]{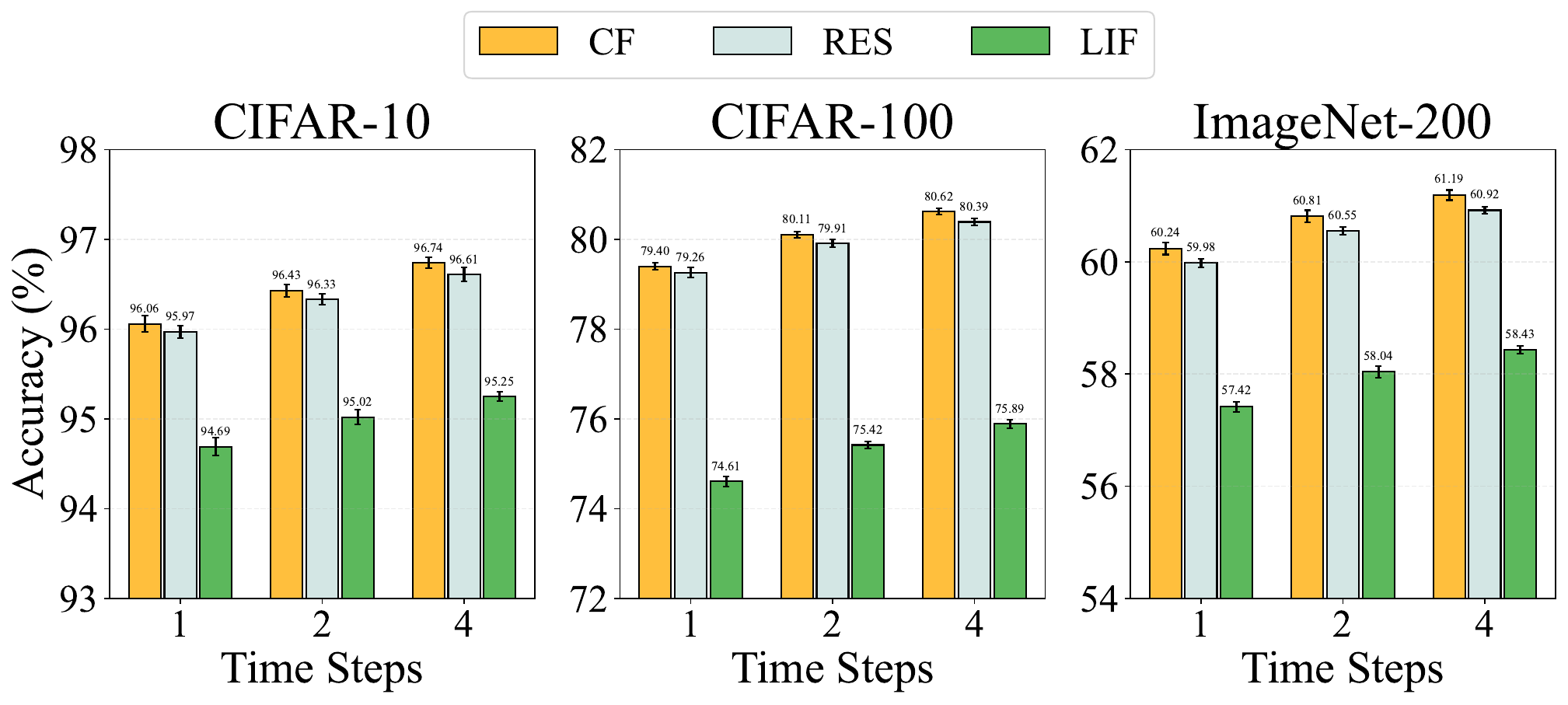}
\caption{Performance comparison of three SNN variants: 
a model employing CF neurons across all spiking layers (CF), 
a baseline model using LIF neurons across all spiking layers (LIF), 
and a hybrid model where CF neurons are introduced only after 
the residual structure (RES).}
\label{residual_bar}
\vspace{-1.0em} 
\end{figure}

\begin{figure}[htbp]
\centering
\includegraphics[width=0.48\textwidth]{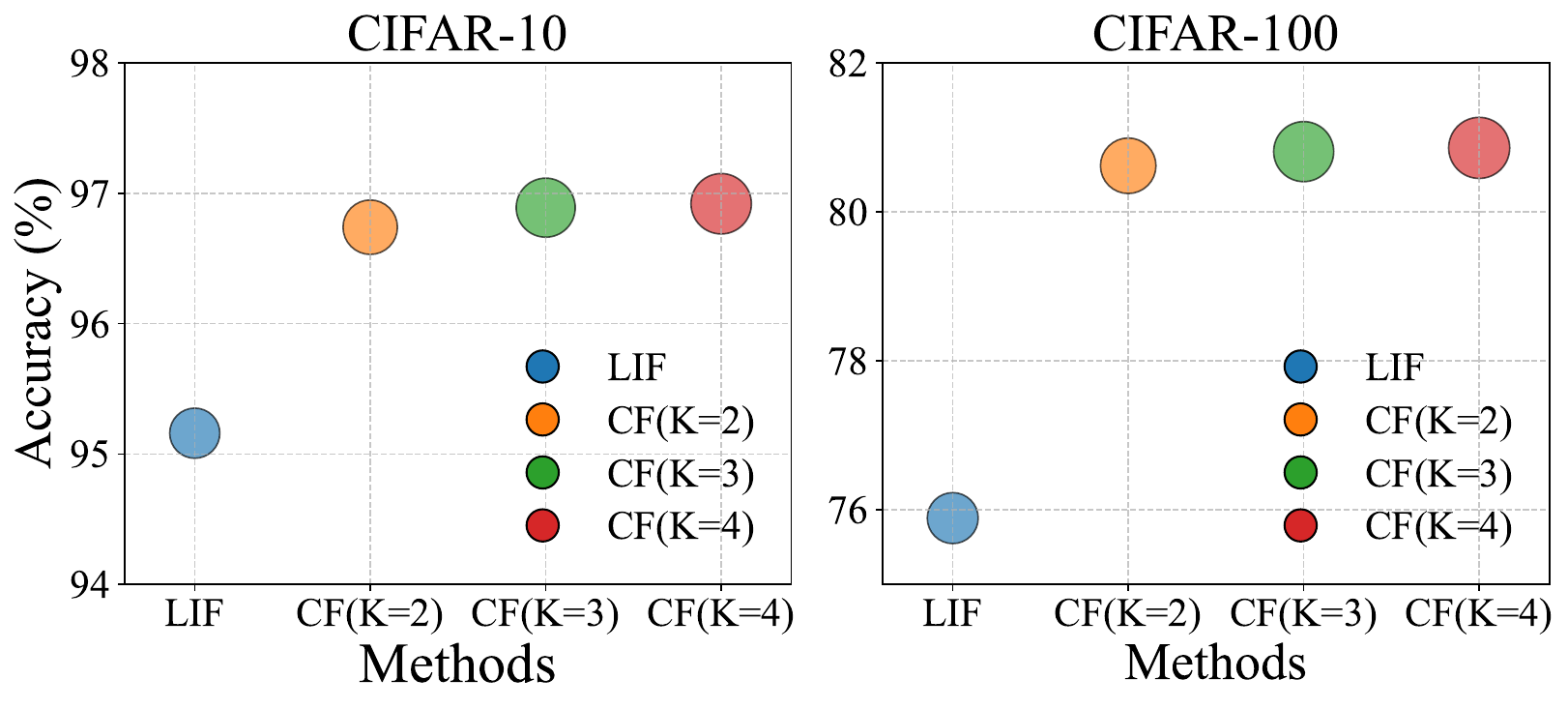}
\caption{Comparison of different methods, where the area of the scatter corresponds to the number of spikes.}
\label{bubble}
\vspace{-1.0em} 
\end{figure}

To further validate that the proposed CF neuron alleviates the 
information loss introduced by residual structures, we compare 
three SNN configurations: 
(i) a model employing CF neurons across all spiking layers (CF SNN), 
(ii) a baseline model using LIF neurons across all spiking layers (LIF SNN), 
and (iii) a hybrid model in which CF neurons are introduced only 
after the residual structure (RES SNN).
As shown in Fig.~\ref{residual_bar}, the RES SNN achieves performance 
close to that of the full CF SNN, while consistently outperforming 
the LIF SNN by a clear margin. Notably, although CF neurons are 
applied only at positions following the residual connections, 
the resulting performance improvement is substantial.
These results indicate that the performance degradation in the 
LIF SNN is closely related to membrane potential information loss 
caused by residual signal aggregation. By enhancing the information 
representation capability at residual structure, the CF neuron 
effectively compensates for such loss, thereby mitigating the 
adverse impact introduced by the residual structure as mentioned above.

Additionally, we conduct experiments to evaluate the influence of the boundary numbers of circulate-firing mechanism. Fig.~\ref{bubble} represents the accuracies and the number of spikes of four methods (vanilla LIF, CF(K=2), CF(K=3), CF(K=4)). 
The results show that increasing 
K leads to more spikes and higher accuracy, but also higher energy consumption. While CF(K=3) and CF(K=4) achieve better performance, they introduce more spikes. In comparison, CF(K=2) provides a better trade-off between performance and energy efficiency. 

\subsection{The Noise Robustness of Our Methods}

\begin{figure}[htbp]
\centering
\includegraphics[width=0.48\textwidth]{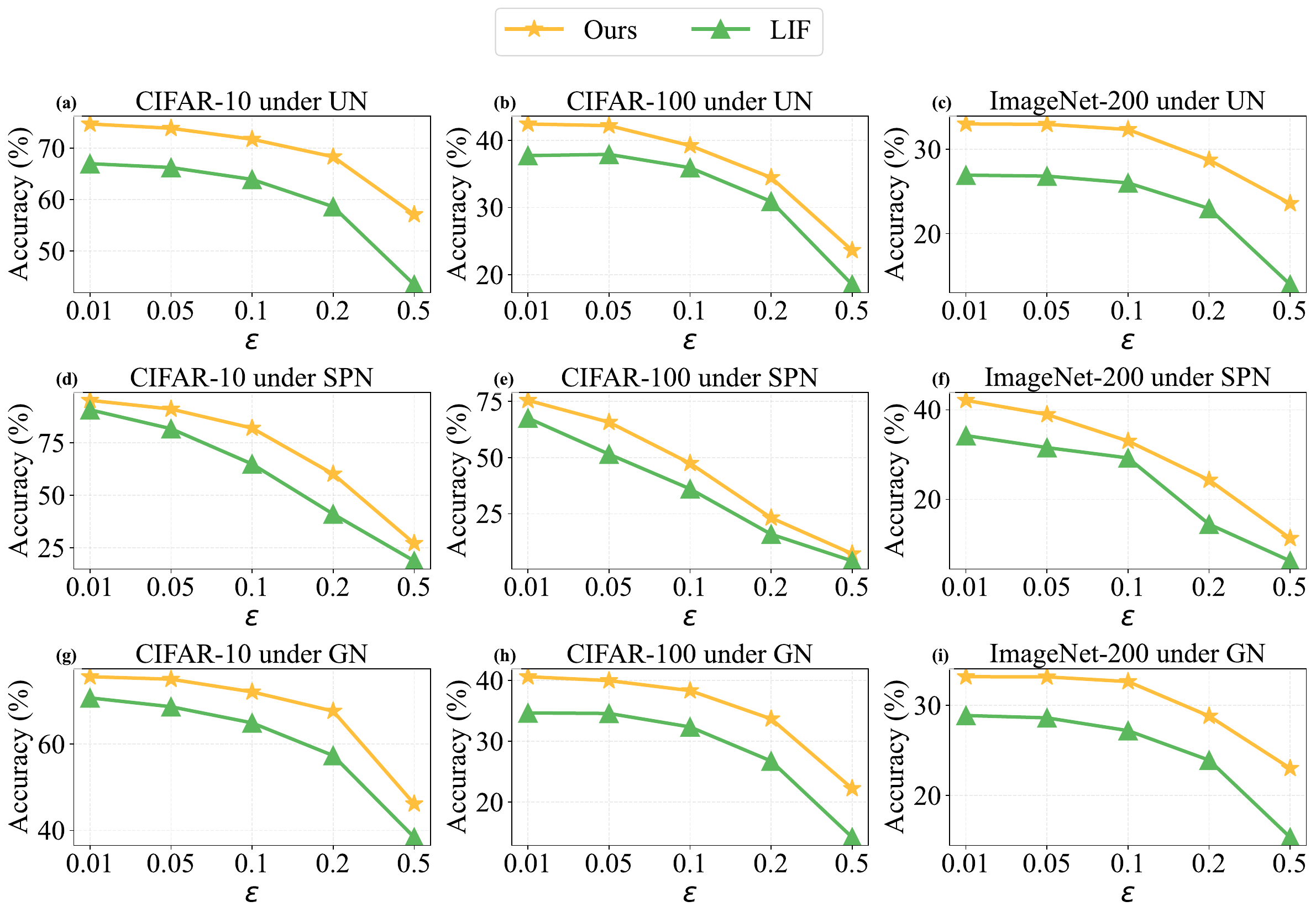}
\caption{The performance of our methods under three noise environments. `UN' represents uniform noise. `SPN' represents salt-and-pepper noise. `GN' represents Gaussian noise.}
\label{noise_plot}
\vspace{-1.0em} 
\end{figure}

To further evaluate the robustness of the proposed method, we conduct comprehensive noise-injection experiments on CIFAR-10, CIFAR-100, and ImageNet-200. 
Three commonly used noise types are considered: (1) \emph{uniform noise}, (2) \emph{salt-and-pepper noise}, and (3) \emph{Gaussian noise}. 

For each noise type, we gradually increase the noise intensity and report the classification accuracy under different noise levels. 
The experimental results are illustrated in Fig.~\ref{noise_plot}, where each row corresponds to one noise type and each column corresponds to one dataset, and $\epsilon$ denotes the noise intensity.
Our method consistently demonstrates strong robustness across all noise types and datasets. 
For uniform noise, the accuracy of our model decreases more slowly compared with other baseline methods, indicating that the proposed neuron model and time-step-wise surrogate gradient effectively enhance the stability of membrane potential dynamics under perturbations. Similar observations retain under Gaussian noise.
For salt-and-pepper noise, which heavily disrupts pixel intensities, our approach still maintains a significant performance advantage, suggesting that the improved firing representation retains discriminative information even in the presence of severe input corruption. 

Overall, the robustness evaluation confirms that our model not only achieves high clean-data accuracy but also generalizes well under noisy conditions. 

\subsection{The Theoretical Analysis of Energy Consumption}

\begin{table*}\normalsize
\centering
\caption{The theoretical energy consumption compared with Transformer-based SNNs, `T' denotes `Time Steps', `Param (M)' denotes `Parameters (Million)'.}
\label{energy}
\begin{tabular}{ccccc}
\toprule
    \hline
    Model & T & Param (M) & SOPs (G) & Power (mJ) \\
    \cmidrule{1-5}
    Spikformer-8-384 \cite{zhou2022spikformer} & 4 & 16.81 & 6.82 & 0.525 \\
    Spikformer-6-512 \cite{zhou2022spikformer} & 4 & 23.37 & 8.69 & 0.669 \\
    Spikformer-8-512 \cite{zhou2022spikformer} & 4 & 29.68 & 11.09 & 0.854 \\
    ResNet-18 (ANN) & $-$ & 11.17 & $-$ & 7 \\ 
    ResNet-18(Ours) & 1, 2, 4 & 11.17 & 1.52, 3.04, 6.08 & 0.117, 0.234, 0.468 \\  
    \hline
    \bottomrule
\end{tabular}
\vspace{-1.0em} 
\end{table*}

To quantitatively evaluate the energy efficiency of different models, we conduct a theoretical energy analysis. The total energy consumption of a neural network can be decomposed into two primary components: the energy consumed by multiply-accumulate (MAC) operations and the energy associated with spike transmissions. 
We employ the approach in \cite{zhou2022spikformer}. 
To estimate the theoretical energy consumption of the network, we first compute the number of synaptic operations (SOPs) for each block or layer $l$ in the SNN model:

\begin{equation}
\text{SOP}_{\text{s}}(l) = f_r \times T \times \text{FLOPs}(l)
\end{equation}

Here, $l$ denotes a specific block or layer, $f_r$ represents the firing rate of the input spike train to that block/layer, and $T$ is the total number of simulation time steps for the spiking neurons. 
FLOPs($l$) corresponds to the floating-point operations required by layer $l$. 
Since energy consumption is proportional to the number of operations, the theoretical energy consumption of block $l$ is computed as:

\begin{equation}
\text{Power}(l) = 77~\text{fJ} \times \text{SOP}_{\text{s}}(l)
\end{equation}
where 77fJ denotes the energy cost per synaptic operation \cite{hu2021spiking, indiveri2015neuromorphic}.

For ANNs, the energy cost is directly related to the number of FLOPs, calculated as:

\begin{equation}
\text{Power}(l) = 12.5~\text{pJ} \times \text{FLOPs}(l)
\end{equation}
where 12.5pJ represents the energy consumption per FLOP. 
Tab.~\ref{energy} represents the theoretical analysis of energy consumption of our methods, other SNN models and ANN model. Our approach consistently achieves lower energy consumption. Notably, under the one-time-step setting, our method significantly reduces energy consumption compared to multi-step SNN models and ANN model, highlighting its suitability for low-latency and low-power scenarios.

\section{Conclusion}
\label{conclusion}

In this work, we propose a direct SNN training algorithm with novel contributions in spiking neuron model, surrogate gradient function and loss function. 
Our proposed approach achieves superior performance compared to existing direct training algorithms under different architectures and closes the performance gap between SNNs and ANNs. 
In addition, our methods demonstrate remarkable generalizability to other direct training algorithms of SNNs under both convolutional neural networks (CNNs) architecture and Transformer architecture. 
Our approach also exhibits enhanced robustness compared to vanilla LIF spiking neuron model. 
The superior performance, architectural compatibility and robustness of our approach enable optimization of existing direct training methods. 
Our approach open a new avenue for bridging the performance gap between SNNs and ANNs. 

\bibliographystyle{IEEEtran}
\bibliography{ref}

\end{document}